\documentclass{article}

\usepackage[final]{neurips_2023}

\usepackage[utf8]{inputenc} 
\usepackage[T1]{fontenc}    
\usepackage{hyperref}       
\usepackage{url}            
\usepackage{booktabs}       
\usepackage{amsfonts}       
\usepackage{nicefrac}       
\usepackage{microtype}      
\usepackage[dvipsnames]{xcolor}      
\usepackage{tcolorbox}
\usepackage{soul}
\usepackage{graphicx}
\usepackage{natbib}
\usepackage{appendix}
\usepackage{todonotes}
\usepackage{amsmath}

\renewcommand{\cite}[1]{\citep{#1}}

\title{Let's Reinforce Step by Step}

\author{%
  Sarah Pan \\
  MIT Primes \\
  Phillips Academy\\
  \texttt{span24@andover.edu} \\
  \And
  Vladislav Lialin  \\
  University of Massachusetts Lowell \\
  \texttt{vlialin@cs.uml.edu} \\
  \And
  Sherin Muckatira \\
  University of Massachusetts Lowell \\
  \texttt{smuckati@cs.uml.edu} \\
  \And
  Anna Rumshisky \\
  University of Massachusetts Lowell \\
  \texttt{arum@cs.uml.edu} \\
}

\begin{document}

\maketitle

\begin{abstract}
  While recent advances have boosted LM proficiency in linguistic benchmarks, LMs consistently struggle to reason correctly on complex tasks like mathematics.
  We turn to Reinforcement Learning from Human Feedback (RLHF) as a method with which to shape model reasoning processes. In particular, we explore two reward schemes, outcome-supervised reward models (ORMs) and process-supervised reward models (PRMs), to optimize for logical reasoning. Our results show that the fine-grained reward provided by PRM-based methods enhances accuracy on simple mathematical reasoning (GSM8K) while, unexpectedly, reducing performance in complex tasks (MATH).
  Furthermore, we show the critical role reward aggregation functions play in model performance.
  Providing promising avenues for future research, our study underscores the need for further exploration into fine-grained reward modeling for more reliable language models.
\end{abstract}

\section{Introduction}

While the linguistic ability of language models (LMs) as measured by NLP benchmarks has increased in recent years, \cite{brown2020language,touvron2023llama} improving their proficiency in complex reasoning tasks like mathematics remains a challenge \cite{yao2023tree}. When prompted with a math problem, LMs will often generate solution steps that seem plausible but are neither correct nor follow logically from previous steps. We believe that the process used to train LMs is partially responsible for these \textit{hallucinatory}, or in other words fallacious, responses \cite{maynez2020faithfulness,uesato2022solving,lightman2023lets,yang2023preferencegrounded}.

``Chain of Thought'' (CoT)  methods, which aim to instill step-wise reasoning at generation-time, increase mathematical performance \cite{wei2023chainofthought}. Simultaneously, CoT prompting does not guarantee a logical response -- little incentive is placed on logic or factual correctness in the original language modeling objective. Reinforcement Learning from Human Feedback (RLHF) \cite{ziegler2020finetuning}, however, allows us to specify the criteria we optimize for, which in this case is logical reasoning. We explore the two reward schemes that have emerged recently -- outcome-supervised reward models (ORMs) and process-supervised reward models (PRMs) \cite{uesato2022solving}. While \citet{lightman2023lets} show that PRMs can identify correct solutions better than ORMs, they did not update a generator model. It is also important to note that our method differs from that of \citet{yang2023preferencegrounded, wu2023finegrained} as we seek not to increase performance directly but rather to improve the logical stepping stones upon which a final answer is generated.

In lieu of using reward models as verifiers \cite{cobbe2021training, lightman2023lets}, we combine CoT and RLHF to guide the internal reasoning process of the LM. We show that accuracy on mathematical benchmarks MATH and GSM8K \cite{hendrycks2021measuring,cobbe2021training}, improves after policy optimization using ORMs and PRMs. Though PRM-based methods lead to decreases in MATH performance, they notably also increase performance on GSM8K by 33\% (relative). Using outcome-supervised reward models does not improve performance on GSM8K but increases accuracy on MATH by 18\%.

\section{Method}
\begin{figure}
    \centering
    \includegraphics[width=0.7\textwidth]{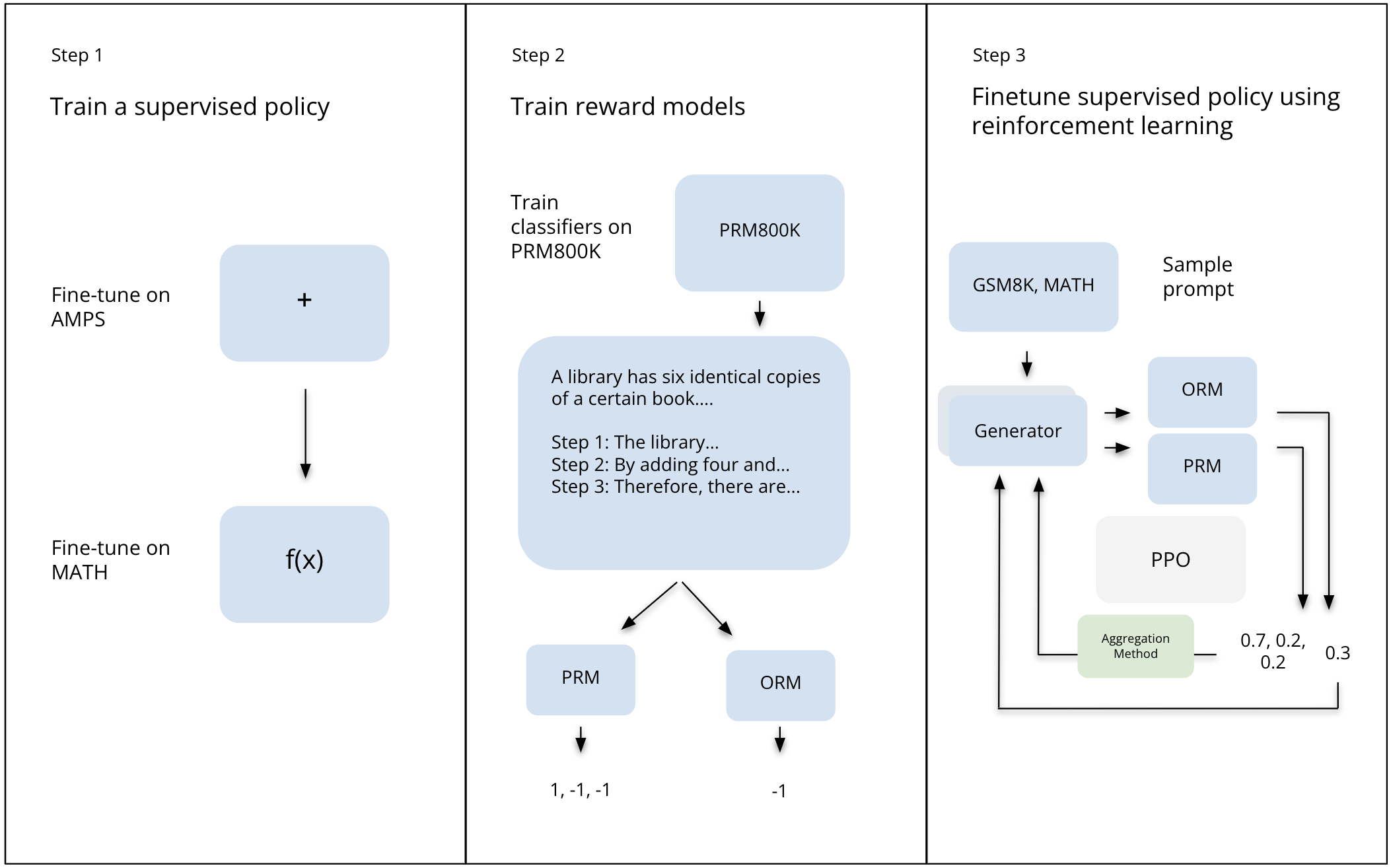}
    \caption{A visual guide to our method, which follows InstructGPT \cite{ouyang2022training}.}
    \label{fig:method-diagram}
\end{figure}

Our overall framework resembles that of \citet{ouyang2022training}. Shown in Figure \ref{fig:method-diagram}, the stages are supervised training, reward model training, and reinforcement learning with the learned reward model.

When performing RLHF, we look at the effects of various PRM reward aggregation schemes. We note that our exploration of aggregation methods is not exhaustive but provides insight into particularly promising methods. We hypothesize that the more fine-grained feedback provided by the PRM provides better signals than that from an ORM. When aggregated effectively, we expect that the PRM improves policy performance on mathematical benchmarks.

\subsection{Supervised Finetuning and Reward Model Training}
We first fine-tune the generator model on multi-step math solutions. Next, we train our reward models. 
To train the outcome-supervised reward model, an $n$-token long question and step-by-step solution are passed in as context.
Only the logits corresponding to the last token are taken into account. \footnote{This is similar to GPT-1 \cite{Radford2018ImprovingLU}, where the hidden state of the last token is used as a representation vector for the entire text.} The label denotes whether that solution contains an incorrect reasoning step.

\begin{equation}
    \centering
    \mathcal{L}_i = \begin{cases}
    - \frac{1}{|S|} y_i \cdot \log p_i & i \in S \\
    0 & i \not \in S
    \end{cases}
    \label{fig:PRM loss objective}
\end{equation}

The process-supervised reward model is trained on a similar classification objective, but as shown in Equation \eqref{fig:PRM loss objective}, there exists a set $S$ that contains indices corresponding to the last token in a reasoning step. All other tokens are masked when calculating the loss.

\subsection{Reward Aggregation Methods}
Using the ORM and PRM trained in phase two, we fine-tune our generator model using PPO \cite{schulman2017proximal}. Our study delves into five distinct reward delivery paradigms: vanilla ORM, PRM-Avg, PRM-Prod, PRM-Max, and PRM-Min. For all the hyperparameters we use, look to Appendix \ref{sec:training details}.

\paragraph{Vanilla ORM}
For the vanilla ORM method, we denote the probability of there being a misstep as $r_\text{ORM}$. While all other tokens receive an estimate of the KL divergence from the original policy as a reward, the last token in the response additionally receives $r_\text{ORM}$. This is shown in Equation \eqref{fig:vanilla reward scheme} for an $n$-token long response.

\begin{equation}
    \centering
    r_t = \begin{cases}
        t \neq n & \text{KL-div}_{t} \\
        t = n & \text{KL-div}_{t} + r_{\text{ORM}}
    \end{cases}
    \label{fig:vanilla reward scheme}
\end{equation}

\paragraph{Aggregated PRMs}
The general equation for reward aggregation is shown by Equation \eqref{fig:base method}. The PRM-Avg method uses $\texttt{aggregate} = \frac{1}{n} \sum_{i = 0}^{n}$ for an $n$-token long sequence. Similarly, the PRM-Prod method uses $\texttt{aggregate} = \prod_{i = 0}^{n}$. 

The PRM-Max approach uses $\texttt{aggregate} = \text{max}$, where the largest per-step reward is used for the entire generation. This method explores the idea that one decisively good reasoning step might be all that is needed to produce a correct final answer. Conversely, PRM-Min uses $\texttt{aggregate} = \text{min}$ and penalizes the model according to the worst reasoning step.

\begin{equation}
    \centering
    r_t = \begin{cases}
        t \neq n & \text{KL-div}_{t} \\
        t = n & \text{KL-div}_{t} + \texttt{aggregate}({\text{PRM}})
    \end{cases}
    \label{fig:base method}
\end{equation}




\section{Experiments}

\subsection{Experimental Setup}
We fine-tune OPT-1.3B \cite{Zhang2022OPTOP} as our generator model. We first train on the Auxiliary Mathematics Problems and Solutions (AMPS) dataset and MATH \cite{hendrycks2021measuring} for one and ten epochs, respectively. Like \citet{lewkowycz2022solving}, we find that supervised fine-tuning is necessary to produce meaningful responses. We evaluate this model as the SFT Base model.

Then, we train two 300M-parameter DeBERTav3\footnote{DeBERTav3 accuracy on the reward modeling dataset was much better than OPT-350M/1.3B in all our experiments.} reward models as a sequence-level classifier and a reasoning-step-level classifier, respectively. Both of these models are trained on the PRM800K dataset \cite{lightman2023lets}, which consists of prompts from MATH \cite{cobbe2021training}, step-by-step responses generated by a model, and step-level labels from human labelers on whether each step is correct.

Following \citet{yao2023deepspeedchat}, we initialize the value function from a reward model with the same tokenizer. We use an OPT-1.3B \cite{Zhang2022OPTOP} model trained on the ORM objective. We choose to train our critic with the ORM objective to avoid the unnecessary complexities of using a PRM. During RLHF, we utilize prompts from both MATH and GSM8K. This provides the generator with greater exposure to math with varying levels of difficulty. Lastly, because PRM800K contains examples from the MATH test set, we filter the overlap to arrive at approximately five hundred questions. We evaluate on the GSM8K test set as is.

\subsection{Results}

  

\begin{figure}
    \centering
    \includegraphics[width=0.8\textwidth]{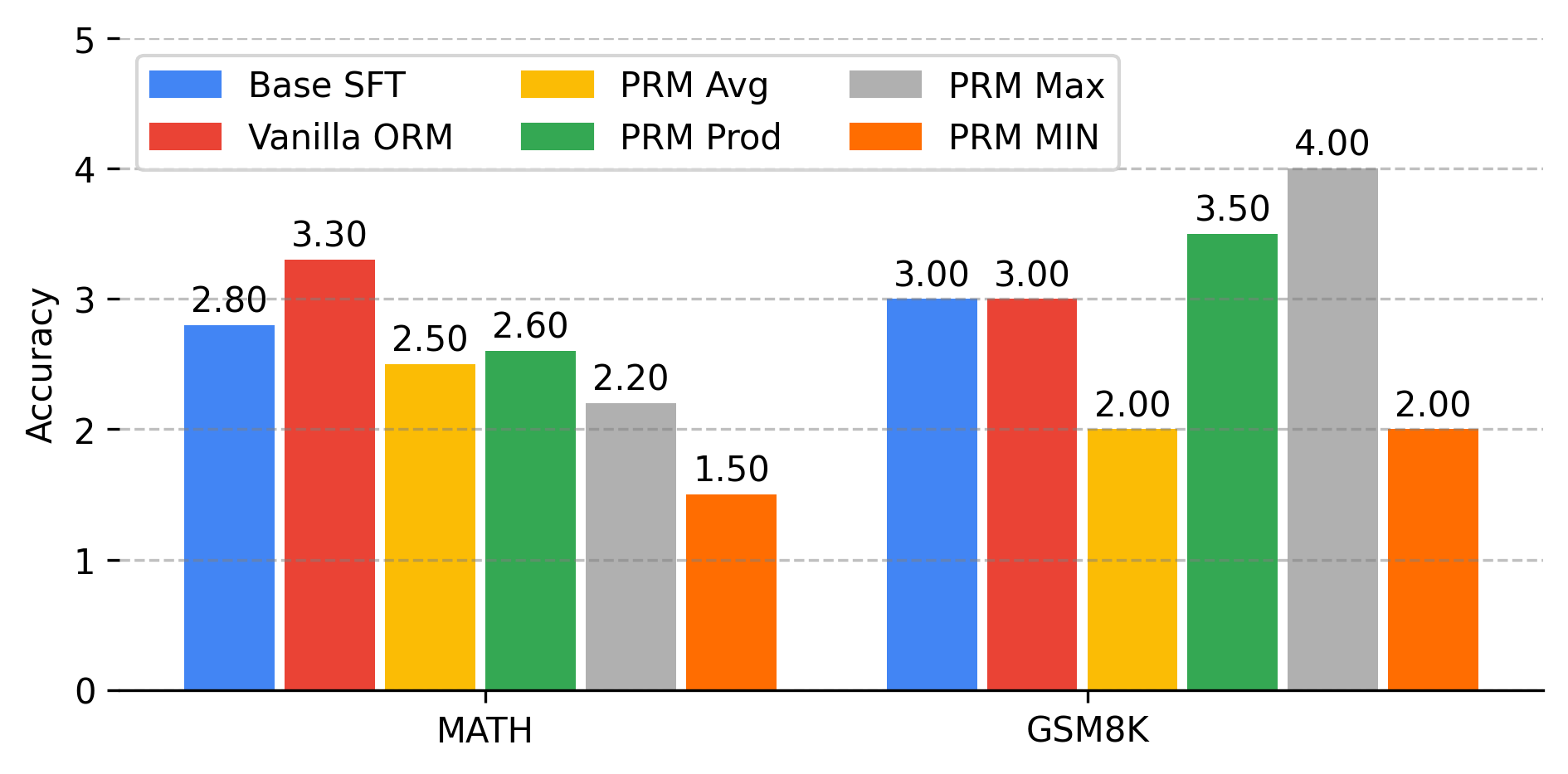}
    \caption{Results on MATH and GSM8K. PRM-based methods degrade performance on MATH but improve accuracy on GSM8K. ORM-based methods improve accuracy on MATH but decrease that on GSM8K.}
    \label{fig:results bar chart}
\end{figure}

Our main results are presented in
Figure \ref{fig:results bar chart}.
We see that providing feedback on LM reasoning process through RLHF is a promising method with which to improve mathematical performance as judged by the accuracy of the final answer (up to 17\% on MATH 33\% on GSM8K, relative).

We find a strong dependence of accuracy on the reward aggregation method. The best performance on GSM8K results from the use of the PRM-Max method, and the best score on MATH comes from the vanilla ORM method. Though PRM-based methods (excluding PRM-Avg and PRM-Min) increase performance on GSM8K, they consistently decrease performance on MATH. The significant decrease in performance after using PRM-Avg and PRM-Min points to the importance of choosing an appropriate aggregation method when using a process-supervised reward model. 

The increase in MATH performance with the outcome-supervised reward model may stem from its holistic training, enabling better handling of complex problems, while its lack of exposure to simpler reasoning hinders decisive signaling for GSM8K problems. In contrast, the process-supervised reward model, also trained on PRM800K, prioritizes correctness on the step level, resulting in improved GSM8K performance but a decline in MATH performance. This hints that the fine-grained knowledge possessed by a PRM may be more akin to ``real math'' and generalize better onto different levels of complexity. Simultaneously, this implies potential limitations in the PRM's ability to effectively understand and reward complex mathematics. \citet{lightman2023lets} use PRMs that are presumably much larger than ours and thus may be better at understanding complex patterns. This study suggests the need for deeper exploration into the trade-offs between ORM and PRM-based methods, especially in terms of varying model scales.

\paragraph{Effects of dataset mixing}
To verify our strategy of mixing GSM8K and MATH training sets in stage threes, we use either dataset with the PRM-Prod method and observe significant decreases in model ability. When only training on MATH, a 0.5\% accuracy on GSM8K and a 2.6\% on MATH is observed. Similarly, only training on GSM8K produces an accuracy of 1.5\% on MATH and 1.5\% on GSM8K. When compared with our results in Figure \ref{fig:results bar chart}, it is clear that training on both datasets is beneficial for performance.

\section{Conclusion}

We have shown that the use of reinforcement learning with outcome-supervised and process-supervised reward can increase performance on complex reasoning tasks like mathematics.
We observe up to 33\% accuracy increase on GSM8K
and a 18\% increase on MATH. Additionally, we find a strong performance dependence on the reward aggregation function for PRM-based methods. For example, maximum-aggregated PRM lead to the best result on GSM8K dataset but resulted in worse-than-baseline performance on MATH.
Moreover, the failure of the average-aggregated PRM highlights the significance of choosing the right aggregation function.
In future work, we plan to explore the use of a non-aggregated step-by-step or, in general, fine-grained rewards directly in the reinforcement learning objective.
Additionally, as \citet{lightman2023lets} found model size to be a crucial factor for step-by-step rewards to be beneficial, we plan to scale up our experiments to larger models and datasets.
We hope that our study encourages more research in fine-grained reward modeling -- especially in relation to the effectiveness of various aggregation methods -- and brings us closer to more reliable language models.


\bibliography{references}

\appendix
\section*{Appendix}

\section{Training Details}
\label{sec:training details}
For all experiments, a weight decay of 0.1 and cosine learning rate scheduler is used. We initialize the generator model from the pre-trained checkpoint of OPT-1.3b. We use a $lr = 6e^{-5}$ to train on AMPS for one epoch using batch size of 152. We further fine-tune this model on MATH for ten epochs using the same batch size.

For the PRM-Avg result, RLHF using prompts from both MATH and GSM8K was performed using batch size 144 and $\text{lr} = 1e^{-4}$ for both actor and critic models. Similar hyperparameters were used for PRM-Prod with a batch size of 126 and $\text{lr} = 1e^{-4}$ for both actor and critic models. The PRM-Max result was achieved using batch size of 160 and $\text{lr} = 1e^{-4}$ for the actor and $\text{lr} = 5e^{-5}$ for the critic. Lastly PRM-Min used batch size 144, $\text{lr} = 1e^{-4}$ for the actor, and $\text{lr} = 5e^{-5}$ for the critic.

During RLHF, we found the coefficient for KL-divergence to have a nontrivial role in model stability. After some experimentation we found that  $\text{kl coeff} = 0.2$ provided optimal results for our setup. Further, we clip our rewards at 0.7, and use 0.2 as our clip range when calculating the loss for both the actor and critic model. Following \citet{ouyang2022training}, we set $\lambda = 0.95$ and $\gamma = 1.0$ in the generalized advantage estimate.


\end{document}